\documentclass[runningheads]{llncs}
\usepackage[T1]{fontenc}
\usepackage{float} 
\usepackage{subfig}
\usepackage{graphicx,wrapfig,lipsum}
\usepackage{amsmath,amssymb,amsfonts}
\newtheorem{assumption}{Assumption}
\usepackage{cite}
\usepackage{url}

\begin{document}

\title{Learning When to Treat Business Processes: Prescriptive Process Monitoring with Causal Inference and Reinforcement Learning}
\titlerunning{Learning When to Treat Business Processes}
%
\author{Zahra Dasht Bozorgi\inst{1} \and
Marlon Dumas\inst{2} \and
Marcello La Rosa\inst{1} \and
Artem~Polyvyanyy\inst{1} \and
Mahmoud Shoush\inst{2} \and
Irene Teinemaa\inst{3}\thanks{Now at DeepMind.}
}
\authorrunning{Z. D. Bozorgi et al.}
%
\institute{University of Melbourne, Parkville VIC 3010, Australia\\
\email{zahra.dashtbozorgi@student.unimelb.edu.au}\\
\email{\{marcello.larosa,artem.polyvyanyy\}@unimelb.edu.au}\and
University of Tartu, Narva mnt 18, 51009 Tartu, Estonia\\
\email{{\{marlon.dumas,mahmoud.shoush\}@ut.ee}}\and
\email{irene.teinemaa@gmail.com}}
\maketitle              
\begin{abstract}
\vspace{-3mm}
Increasing the success rate of a process, i.e. the percentage of cases that end in a positive outcome, is a recurrent process improvement goal. At runtime, there are often certain actions (a.k.a.\  treatments) that workers may execute to lift the probability that a case ends in a positive outcome. For example, in a loan origination process, a possible treatment is to issue multiple loan offers to increase the probability that the customer takes a loan. Each treatment has a cost. Thus, when defining policies for prescribing treatments to cases, managers need to consider the net gain of the treatments. Also, the effect of a treatment varies over time: treating a case earlier may be more effective than later in a case. This paper presents a prescriptive monitoring method that automates this decision-making task. The method combines causal inference and reinforcement learning to learn treatment policies that maximize the net gain. The method leverages a conformal prediction technique to speed up the convergence of the reinforcement learning mechanism by separating cases that are likely to end up in a positive or negative outcome, from uncertain cases. An evaluation on two real-life datasets shows that the proposed method outperforms a state-of-the-art baseline.

\end{abstract}
\section{Introduction}
Prescriptive process monitoring is a family of techniques to recommend actions (herein called \emph{treatments}) that, if executed, are likely to optimise a process with respect to one or more process performance indicators~\cite{kubrak2022prescriptive}. For example, in a loan origination process, a treatment could be to send an additional loan offer with better conditions to a customer who is hesitating to accept a loan. This treatment is intended to increase a performance indicator known as the success rate -- the percentage of cases that end in a positive outcome, which in this context means ending in an accepted loan offer.

Each treatment has a cost. This cost is often sufficiently high to make it impractical to treat every case. Furthermore, the effect of the treatment might be different across cases. A treatment that works well for one case, might be ineffective on others. Another important aspect of applying treatments is their timing. Coming back to the loan origination example, sending an additional offer later in the process might be less effective than sending it earlier. 

Previous studies on prescriptive process monitoring propose to produce treatment recommendations by using machine learning models --  trained on historical execution data -- to predict the outcome of each case~\cite{alarmBased, DBLP:journals/kais/Fahrenkrog-Petersen22}. In particular, recommending treatments using online reinforcement learning (RL), combined with predictive models, has shown promising results~\cite{metzger2020triggering}. However, this prior approach has two key limitations. First, given that it makes recommendations based on outcome predictions, it tends to treat cases that are likely to end up in a negative outcome, even when treating a case is unlikely to switch its outcome from negative to positive~\cite{DBLP:conf/icpm/BozorgiTDRP21}. In other words, this prior approach does not consider the effectiveness of the treatments. The second limitation is concerned with the use of online RL, which requires learning through trial and error. This means that the RL agent makes mistakes until it eventually learns to perform well. In addition, the convergence of the agent may be slow, e.g.\ the RL agent may need to see hundreds of cases before converging to a satisfactory treatment policy. 

This paper proposes an RL method for learning treatment policies for prescriptive process monitoring, which addresses the above limitations as follows:

\begin{itemize}
    \item To take into account the effectiveness of the treatments, it incorporates causal effect estimations into the RL process.
    \item To train the RL agent offline, it enhances the available dataset with so-called \emph{alternative outcomes}. The enhanced dataset simulates a realistic environment, so that the RL agent can get feedback on its choices offline.
    \item To speed up the convergence of the RL agent, it leverages a method called conformal prediction -- a predictive modeling method that segregates cases that are almost certain to finish in a positive class, from uncertain cases. Armed with this information, the RL agent is able to avoid treating cases that most likely will end up in a positive outcome anyway.
\end{itemize}


The rest of this paper is organised as follows. Section~\ref{sec: priorwork} reviews related work. Section~\ref{sec: preliminaries} introduces relevant concepts and notations. The proposed method is discussed in Section~\ref{sec: method}, while an experimental evaluation is reported in Section~\ref{sec: results}. Finally, Section~\ref{sec: conclusion} draws conclusion and discusses future work.

\section{Related Work}
\label{sec: priorwork}
\textbf{Prescriptive Process Monitoring:}
Many works address the prescriptive monitoring problem. Here, we mention the studies closely related to our work. For a complete review of prescriptive monitoring methods, we refer the reader to~\cite{kubrak2022prescriptive}. 

Teinemaa~\emph{et~al}.~\cite{alarmBased} propose a prediction-based system that uses empirical thresholding to fire alarms when treatment is needed. This work was later extended by Fahrenkrog-Petersen~\emph{et~al}.~\cite{DBLP:journals/kais/Fahrenkrog-Petersen22}, who discovered that firing the treatment later than when the threshold is reached improves the outcome at lower costs. In the work by Metzger~\emph{et~al}.~\cite{metzger2020triggering}, the authors use online reinforcement learning to learn the best time for triggering treatments. They show that RL outperforms empirical thresholding. These methods, however, do not address the effectiveness of the treatment or make simplistic assumptions about effectiveness. In this paper, we aim to address these limitations.

In another line of work, prescriptive methods are used to recommend the next best activity. de Leoni~\emph{et~al}.~\cite{de2020design} prescribes the next task of the process workers helping clients with a job search. Weinzierl~\emph{et~al}.~\cite{10.1007/978-3-030-58638-6_12} prescribe the next activity which is predicted to maximize the chance of a positive outcome. More recently, Padella~\emph{et~al}.~\cite{padella2022explainable} provided an explainability framework for prescriptive analytics of business processes. These works differ from ours in that the next best action is not a special treatment, only applied when the case is in a negative state. But the actions are part of the normal process execution.

\textbf{Causal Process Mining:}
Causal process mining is an emerging sub-field in process mining concerned with the discovery and estimation of causal effects in processes. Koorn~\emph{et~al}.~\cite{DBLP:conf/bpm/KoornLLR20} discover the cause-effect relationship between a worker's response to aggressive situations and its effectiveness. They account for possible confounding variables in~\cite{koorn2022mining}. Some approaches use structural equation models to discover root causes and answer counterfactual questions about undesired outcomes\cite{DBLP:conf/bpm/QafariA20, qafari2021case}. These methods are concerned with the discovery of causal effects, while we use causal effect estimation for outcome improvement.
Our previous work, we proposes a rule-based recommendation method based on causal effects to improve process outcomes~\cite{DBLP:conf/icpm/BozorgiTDRP20}. In another work, we use causal effects to address process duration reduction at runtime~\cite{DBLP:conf/icpm/BozorgiTDRP21}. While our latter works focus on finding the best cases to treat, in this work, we focus on when to treat a case.


\section{Background}
\label{sec: preliminaries}
This section introduces process mining-related definitions used in the rest of the paper and introduces causal inference, reinforcement learning, and conformal prediction concepts upon which the proposal relies.

\subsection{Process Mining}
\begin{definition}[Event, Trace, Event Log]
\label{def:trace event log}
\emph{An \emph{event} is a tuple $(a,c,t,(d_1, v_1), \\ \ldots,(d_m, v_m))$, $m \in \mathbb{N}_0$, where 
$a$ is an activity name, 
$c$ is a case identifier, 
$t$ is a timestamp, and 
$(d_1, v_1), \ldots, (d_m, v_m)$ are attribute-value pairs.
A \emph{trace} is a finite sequence $\sigma = \langle e_1,\ldots,e_{n} \rangle$, $n \in \mathbb{N}$, of events with the same case identifier in ascending timestamp order. 
An \emph{event log}, or \emph{log}, is a multiset of traces.}
\end{definition}

\begin{definition}[k-Prefix]
\label{def: prefix}
\emph{A \emph{k-prefix} of a trace $\langle e_1,\ldots,e_{n} \rangle$, $n \in \mathbb{N}_0$,
is a sequence $\langle e_1,\dots, e_{k} \rangle$, $0 \leq k \leq n$.}
\end{definition}

\begin{definition}[Sequence encoder]
\label{def: seqEncoder}
\emph{A sequence encoder $f:S \rightarrow X_1 \times \cdots \times X_p$ is a function that takes a (partial) trace $\sigma$ and transforms it to a feature vector $X$ in the p-dimensional vector space $X_1 \times \cdots \times X_p$ with $X_i \subseteq \mathbb{R}, 1 \leq i \leq p$.}
\end{definition}

\subsection{Causal Inference}
We use the \textit{Neyman-Rubin Potential Outcomes Framework}~\cite{rubin1974estimating} for causal inference. An \textit{intervention}, or a \textbf{\textit{treatment}} is an action that can be done during the execution of a process to optimise the outcome of the case. In this paper, we consider the binary treatment setting where the treatment is denoted by a binary variable $T \in \{0,1\}$, with $T=1$ denoting when the treatment is applied to a case and $T=0$ otherwise. According to the potential outcomes framework, each case has two potential outcomes: $Y(1)$ denoting the outcome under treatment and $Y(0)$ the outcome under no treatment. The effectiveness of the treatment is not constant across cases and throughout different points in the case. To measure the effectiveness of the treatment, we use the \textit{Conditional Average Treatment Effect} (CATE):

\begin{definition}[Conditional Average Treatment Effect]
\emph{Let $X$ be a set of attributes that characterize a case.
Then, the \emph{conditional average treatment effect} (CATE) of the case is defined as follows:
\vspace{-1mm}
\begin{center}
$\mathit{CATE}: \theta(x) = \mathbb{E}[Y(1)-Y(0) \,|\, X=x]$.
\end{center}
}
\end{definition}

To identify causal effects, we make the four standard assumptions in the potential outcomes framework: positivity, ignorability, consistency, and no-interference.

\begin{assumption}
    (Positivity). \emph{Every case has the potential to be selected for treatment, that is, $P(T=t \,|\, X=x)>0$, for every treatment $t$ and every vector $x$ representing a prefix.}
\end{assumption}

\begin{assumption}
    (Ignorability). \emph{Treatment assignment is independent of the potential outcomes conditioned on pre-treatment confounders X:} $Y(1),Y(0) \perp \!\!\! \perp T\,|\,X$.
\end{assumption}

\begin{assumption}
    (Consistency). \emph{Observations of outcome after treatment selection are consistent with potential outcomes: $Y=Y(t)$ if $T=t$ for all $t$. }
\end{assumption}

\begin{assumption}
    (No-interference). \emph{Treatment decision for one case does not affect treatment decisions for other cases.}
\end{assumption}

Many approaches for estimating the CATE have been proposed in the literature. One such method that we use in this work is Causal Forest.

\smallskip\noindent\textbf{Causal Forest} is a causal estimator proposed by Athey~\emph{et~al}.~\cite{wager2018estimation}. It is an ensemble of causal trees~\cite{athey2016recursive}. Causal trees are a modification of decision trees that make them suitable for estimating causal effects. In decision trees, the splits aim to separate classes, but in causal trees, they aim to increase the expected causal effect.
One major difference that distinguishes causal trees from decision trees is \textit{honest splitting}. It means that during training, the data is divided into two sets: one for building the tree and the other for the estimation of the treatment effect after the split. A Causal Forest is constructed by aggregating the results of many honest causal trees using the subsampling method. 

\begin{wrapfigure}{r}{0.5\textwidth}
	\centering
    \vspace{-7mm}
	\includegraphics[width=\linewidth]{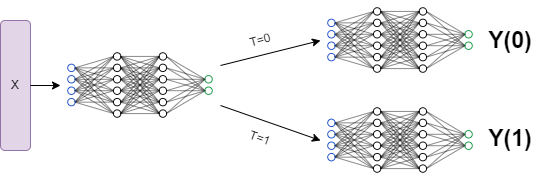}
	\caption{Realcause potential outcome generation architecture}
	\label{fig:realcause}
	\vspace{-6mm}
\end{wrapfigure}

\smallskip\noindent\textbf{Realcause} is a method for generating realistic data with two potential outcomes. In this paper we keep the prefixes real and only generate two potential outcomes. Figure~\ref{fig:realcause} describes the architecture of Realcause. First, a neural network is trained using all of the data (i.e., prefixes) to get a hidden representation. Then samples under treatment and no treatment are separated and two neural networks are trained for each group. The activation functions of the latter neural networks are used to parameterise distributions, which can then be sampled to generate both potential outcomes for each input sample.

\subsection{Reinforcement Learning}
\label{RL}
Reinforcement learning is a branch of machine learning to train models that make a sequence of decisions. Typically, an agent is placed in an environment. It observes a state $s$ and performs an action $a$ from a set of actions in the action space, then observes a reward $r$ and the next state $s'$ as a response to $a$. The goal of the agent is to maximise the total cumulative reward. The agent typically starts by randomly selecting actions and receiving rewards for those actions. Through trial and error, it will figure out which actions in which conditions produce more reward. One of the main problems that the agent faces while learning is finding a good trade-off between exploration and exploitation. Exploitation means that the agent uses the knowledge already acquired to perform actions that are guaranteed to produce a lot of reward. Exploration means that the agent tries new actions to find other potentially rewarding actions. A good balance between exploration and exploitation is essential for the agent to discover all the rewarding actions.

\subsection{Conformal Prediction}
\label{conformalPred}
Conformal prediction is a method to generate prediction sets for any prediction model~\cite{romano2020classification}. Given an uncertainty score by a prediction model, conformal prediction outputs a set of classes that covers the true class with mathematical guarantees. Below we describe how to construct these prediction sets and what guarantees they will have. 

Suppose we have trained a prediction model $\hat{f}$ that outputs probabilities for each prediction class. We use a small set of unseen calibration data of size $n$ to construct \textit{conformal scores} $s_i = 1 - \hat{f}(x_i)_{Y_i}$ where $\hat{f}(x_i)_{Y_i}$ is the predicted probability of the true class. Score $s_i$ is high if the model is very wrong, meaning that $i$ does not conform to the training data. Next, given $s_1, \ldots, s_n$ we define $\hat{q} = \lceil (n+1)(1-\alpha)\rceil/n$ where $\lceil.\rceil$ is the ceiling function and $\alpha$ is a user-defined error tolerance threshold. Finally, using a separate test set, we define the conformal prediction set as $C(X_{test}) = \{y: s(X_{test}) \leq \hat{q} \}$. In~\cite{romano2020classification}, it is shown that the prediction set is guaranteed to contain the true class with probability $1-\alpha$.

\section{Methodology}
\label{sec: method}

\begin{figure*}[t]
\vspace{-5mm}
	\centering
	\includegraphics[width=1\linewidth]{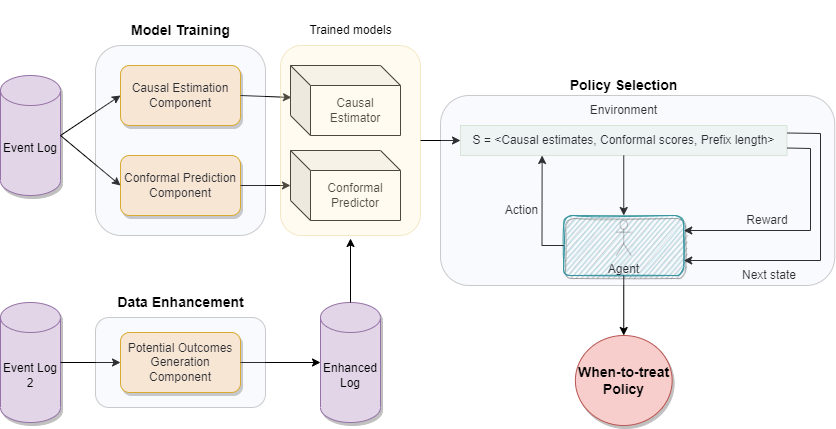}
	\caption{Overview of the proposed approach}
	\label{fig:framework}
	\vspace{-5mm}
\end{figure*}

Suppose agent $\mathcal{A}$ is responsible for making decisions about treating ongoing cases. $\mathcal{A}$'s job is to consider each case after each event and decide whether to treat that case or not. At each decision point (after each event), $\mathcal{A}$ needs to answer the following two questions: Will applying the treatment now change the outcome of the case from negative to positive? How confident am I about the outcome of the case, regardless of treatment? To help $\mathcal{A}$ answer these questions, we train two machine learning models using past process executions: a predictive model estimating probabilities of each possible outcome and a causal model estimating the CATE of the chosen treatment. This is the first phase of the approach. In the second phase, we create a realistic environment for $\mathcal{A}$ to try different treatment policies and learn the best one. To do that, we use a generative model to generate potential outcomes for each prefix length in such a way that they are statistically indistinguishable from the actual outcomes. Finally, in the third phase, we let $\mathcal{A}$ learn the best treatment policy through trial and error. We design a reward function to guide $\mathcal{A}$ about when it makes correct or incorrect decisions. Figure~\ref{fig:framework} provides an overview of our approach. We explain each of the phases below.

\subsection{Model Training}

\subsubsection{Causal Effect Estimation}
\begin{wrapfigure}{r}{0.5\textwidth}

	\centering
    \vspace{-7mm}
	\includegraphics[width=\linewidth]{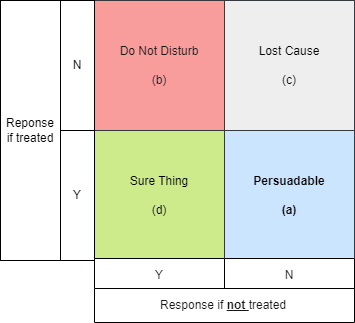}
	\caption{Grouping of cases according to their response to treatment}
	\label{fig:uplift}
	\vspace{-6mm}
\end{wrapfigure}

To answer $\mathcal{A}$'s first question, we train a causal estimator. Different from prediction methods that seek to estimate $P(Y|X)$, causal estimators estimate $CATE: P(Y(1)-Y(0)|X)$. Typically in the causal estimation literature, cases are divided into four groups: (a) Persuadables, (b) Do Not Disturbs, (c) Lost Causes, (d) Sure Things (See Figure~\ref{fig:uplift}). Using causal estimation, we can separate cases into these groups. $CATE>0$ describes the persuadables, $CATE<0$ the do not disturbs, and $CATE=0$ the sure things and the lost causes. 
Since causal estimators are difficult to evaluate, a point estimate of the CATE may not be reliable enough to base treatment decisions on. Instead, we train the model to compute confidence intervals for CATE. The model takes a prefix $\sigma_k$ with $k\leq n$ where $n$ is the total length of the case and returns $\theta_{u,k}$ and $\theta_{l,k}$ which are the upper and lower bound for the estimated causal effect, respectively, for a prefix of length $k$. We use Causal Forest~\cite{wager2018estimation} to get these confidence intervals. Causal Forest and orthogonal random forest (ORF)~\cite{DBLP:conf/icml/OprescuSW19} are two causal estimators that can produce valid confidence intervals. We chose causal forest as we have found that it is quicker to train while having similar performance to ORF when both treatment and outcome are binary. But generally, our framework is independent of the chosen causal estimator, as long as it can produce confidence intervals. 

\subsubsection{Conformal Prediction}
Having a CATE confidence interval allows agent $\mathcal{A}$ to converge to a good policy. Still, there can exist prefixes that the causal estimator is unsure about. For example, when $\theta_{u,k}>0$ and $\theta_{l,k}<0$. So we hypothesise that if we add an additional rigorous notion of uncertainty, such as conformal prediction, we can help the agent converge faster. Recall that with conformal prediction, we can construct prediction sets that are mathematically guaranteed to contain the true class. For binary outcome prediction, these sets will be one of the following: (a) $\{\}$, (b) $\{0,1\}$, (c) $\{0\}$, and (d) $\{1\}$. In (a) and (b), the model is unsure about the outcome. But in (c) and (d), we have a rigorous guarantee that the case will end up with a negative and positive outcome, respectively. These correspond to the lost causes and the sure things. Because conformal prediction tells us that we are confident of the outcome based on the information in the prefix. So, showing these prediction sets to agent $\mathcal{A}$ should help it prune the search space for prefixes that need treatment because it will be sure not to treat the lost causes and the sure things that conformal prediction has identified.

To use conformal prediction, we need to train a predictive model first. This model is a function learned from the data that takes a prefix $\sigma_k$ and returns two probabilities: $p_k(1)$, the probability of a positive outcome at prefix length $k$, and $p_k(0)$, the probability of a negative outcome. The predicted class is defined as the class with the highest probability:

\begin{equation}
    p(\sigma_k)=Argmax(p_k(1), p_k(0))
\end{equation}
We use Catboost~\cite{dorogush2017catboost} to train the predictive model as it has been shown to perform well in recent works~\cite{padella2022explainable}. We then apply the conformal prediction algorithm as described in Section~\ref{conformalPred} to get the prediction sets. Then, we encode the prediction sets as the confidence measure $\rho_k$ that the case will have a positive outcome. If the prediction set is (c) then $\rho_k=0$, if the prediction set is (d) the $\rho_k=1$, and if the prediction set is either (a) and (b) then $\rho_k=0.5$.


\subsection{Data Enhancement}
In online RL, agent $\mathcal{A}$ must try different actions in different states to learn the optimal policy. But since we seek to find this policy using data, we need to simulate the environment in which the agent learns. This environment needs to be realistic enough to reflect the information in the data.  In this phase, we seek to generate alternative outcomes that $\mathcal{A}$ can later use during learning to evaluate its knowledge and make decisions accordingly. 

\begin{wrapfigure}{r}{0.5\textwidth}
	\centering
	\includegraphics[width=\linewidth]{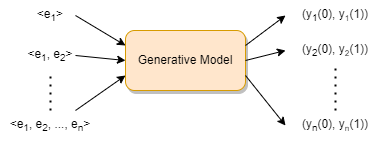}
	\caption{Potential outcomes generation using generative machine learning}
	\label{fig:counterfactuals}
	\vspace{-5mm}
\end{wrapfigure}

In real-world data, we only observe one of the potential outcomes in the Neyman-Rubin framework. Therefore, we cannot know the true unit-level causal effect. Recently, one solution has been to use generative machine learning to produce realistic data that is statistically indistinguishable from the original data. We can then use these generative models to get two outcomes for the same feature vector describing a prefix: one outcome if $T=1$ and another if $T=0$. One such method is called Realcause~\cite{DBLP:journals/corr/abs-2011-15007} which uses a generative machine learning model to generate realistic data including the true causal effect. In our previous work~\cite{journalExtension} we successfully applied the Realcause method to generate encoded feature vectors describing cases. In this paper, however, we would like to keep agent $\mathcal{A}$'s environment as realistic as possible. Thus, we use Realcause to only generate potential outcomes for each prefix of each case (See Figure~\ref{fig:counterfactuals}). 


Realcause is originally designed to handle independent data points. But in our method, each data point is a prefix and prefixes are parts of cases. So, to guide the generative model to discover this dependency, we take the following steps: we include the prefix number as a feature where for each case, this feature describes $k$ for each prefix $\sigma_k$. Next, we convert the case identifiers to numeric values. We then include the numeric case ID and prefix numbers in model training. Because Realcause directly learns the data-generating distribution, it will discover the relationship between the prefix numbers $1 \ldots n$ having the same numeric case ID. We use two multi-layer perceptrons with two hidden layers to model the treated and untreated groups. We assume a Bernoulli distribution to model the process outcome.
Once the generative model is trained on the original event log, we can generate an enhanced event log by sampling the outcome model under both treatment and no-treatment conditions, giving us both potential outcomes for each prefix. We use the the enhanced version of the log in the policy selection phase.

\subsection{Policy Selection}
\label{policyselection}
With two potential outcomes to create the learning environment for $\mathcal{A}$, the next step is to let $\mathcal{A}$ map its knowledge into actions by learning a treatment policy.
A policy is a function $\pi$ that maps a prefix $\sigma_k$ to an action $t$. Our goal is to find a policy that maximises a net gain function. To find such a policy we use reinforcement learning. Specifically, we use the \textit{policy-based} RL framework proposed in~\cite{metzger2020triggering}. Below, we formulate the learning problem using RL.

As mentioned in Section~\ref{RL}, in reinforcement learning, an agent is placed in an environment and observes states $s_k$ which describes the environment at each prefix length $k$. The agent selects an action $t$ from action space $\mathcal{T}$ and observes: the reward $r$ and the next state $s_{k+1}$. The agent then learns the best behaviour by trial and error until it reaches the optimal policy for selecting actions. To translate this learning problem into prescriptive monitoring, we modify the RL-based approach proposed by Metzger~\emph{et~al}.~\cite{metzger2020triggering}. Similar to them, we define a binary action space $\{0,1\}$ with $t=1$ as applying the treatment and $t=0$ as not applying it. Metzger~\emph{et~al}. describe the state $s$ to the agent as a tuple $s=(\delta_k, \gamma_k, k)$, where $\delta_k$ is the predicted deviation from a positive outcome, $\gamma_k$ is a reliability score for the prediction, and $k$ is the prefix length. We propose to modify $s$ to contain the estimated CATE interval and the conformal prediction score: $s=(\theta_{u,k}, \theta_{l,k}, \rho_k, k)$.

One important aspect of the learning problem is defining a suitable reward function. The goal of agent $\mathcal{A}$ in RL is to maximise cumulative rewards. Since we consider the best policy to be one that maximises net gain, the most straightforward reward is the gain or loss that we get at the end of each case.  
So, we have decided to incorporate the treatment cost and the benefit of a positive case into the reward function. The intuition behind this design choice is that the treatment policy is directly related to the ratio between this cost and benefit. For example, if we have a cheap treatment and a high benefit, we can afford to apply the treatment more frequently, even if we are not certain about its effectiveness. But with an expensive treatment, it becomes more important to carefully select the cases and times of treatment, and only treat if the agent is sure that the treatment is necessary and effective. 

The reward function also needs to contain information about the effectiveness of the treatment at each decision point. Recall that in the data enhancement phase, we generated $Y(1)$ and $Y(0)$ for each prefix length of each case. We can compute $Y(1)-Y(0)$ for each prefix to obtain the true treatment effect at each decision point. We include this true effect in the reward function to guide the agent about the effectiveness of the treatment. We provide the details of the reward function in Table~\ref{table:rewards}. 

\begin{table}[]
\centering
\vspace{-5mm}
\begin{tabular}{|c|ccc|}
\hline
                  & \multicolumn{3}{c|}{True Treatment Effect}                                                                                                                   \\ \hline
Agent's Treatment & \multicolumn{1}{c|}{Positive}    & \multicolumn{1}{c|}{Negative}   & Zero                                                                                      \\ \hline
Yes               & \multicolumn{1}{c|}{Gain - Cost} & \multicolumn{1}{c|}{-Cost-Gain} & \begin{tabular}[c]{@{}c@{}}Negative Outcome: -Cost\\ Positive Outcome: -Cost\end{tabular} \\ \hline
No                & \multicolumn{1}{c|}{-Gain}       & \multicolumn{1}{c|}{Gain}       & \begin{tabular}[c]{@{}c@{}}Negative Outcome: 0\\ Positive Outcome: Gain\end{tabular}      \\ \hline
\end{tabular}
\vspace{2mm}
\caption{The proposed reward function}
\label{table:rewards}
\vspace{-10mm}
\end{table}

When the agent treats, if the treatment effect is positive, we give a reward of $r=Gain - Cost$, since we receive the gain of a positive outcome while paying the cost of treating. If the treatment effect is zero, we penalise the agent by giving it a negative reward of $r=-Cost$, since the agent wasted the cost of the treatment and it was ineffective. If the treatment effect is negative, we penalise the agent by giving it an even lower negative reward $r= -Cost - Gain$. We chose this reward because not only did the agent waste the cost of the treatment, but also caused further damage by treating when it hurt the outcome. 

When the agent does not treat, if the treatment effect was positive, we penalise it by giving the reward $r=-Gain$ because the agent failed to act when necessary. If the treatment effect is zero, we look at the outcome. If the outcome is positive, the agent correctly decided not to apply the treatment and saved the cost of treatment, so $r=Gain$. If the outcome is negative, it means that the case is a lost cause and even treating it would not have changed anything. So, although the agent was correct in not treating, we give it $r=0$ because there was no gain. Finally, if the treatment effect was negative, we give it a strong positive reward $r=Gain$ because choosing not to treat when the treatment hurt the outcome caused a positive outcome.

This reward function closely models the net gain achieved by following the agent's policy, except in three situations. First, if the agent does not treat when it would have been effective, the actual gain is $0$, but we penalise the agent with $-Gain$. Second, if the agent treats when its effect is negative, we lose the cost of the treatment, so net gain is $-Cost$, but the reward is $-Cost-Gain$. Third, if the agent treats, and the treatment is ineffective, and outcome is positive, the net gain is $Gain-Cost$, but we give $r=-Cost$. We added these extra \textit{punishments} to signal to the agent that it made incorrect decisions. According to~\cite{sutton2018reinforcement}, reward functions often need to be tweaked to speed up learning and convergence, and to avoid getting stuck in local optima. We found these further punishments are necessary to speed up the learning process.

We use a separate set of cases for the policy selection phase with timeframes later than the cases used for model training. The agent learns through a series of episodes. Each episode corresponds to one case. During each episode, the events are presented to the agent in the ascending order of their timestamps. At the end of each episode, we reveal the cumulative reward to the agent. Similar to~\cite{metzger2020triggering}, we use proximal policy optimization (PPO)~\cite{schulman2017proximal} as our RL algorithm. We also represent the policy as a multi-layer perceptron. This network can be used later as a starting point in real-life situations. Since it has been trained on data with potential outcomes that are statistically similar to their real counterparts, it will make fewer mistakes than if the agent starts from scratch.

\section{Results}
\label{sec: results}

\noindent In this section, we explain our experimental setup and report our results. Our method was developed in Python 3.8. We used the Catboost library for our predictive model and EconML for the causal forest. The generative model was developed using Pytorch. All data splitting has been done temporally, ensuring that the cases in the training set whose timeframe overlaps the timeframe of the test set are removed. We split the data into two halves. The first half is used for training the predictive and causal models, and the second is used for policy selection using reinforcement learning. 
We follow the same pre-processing and feature engineering steps for the predictive, causal, and generative models. We one-hot encode the categorical attributes. For encoding timestamp information, we create the following temporal features: `time since case start', `time since last event', `time since midnight', `month', `weekday', and `hour'. Also, to capture the temporal relationship between cases, we create a feature `time since first case', a case attribute denoting the distance between the start of the case and the start of the first case. To pre-process the data, we standardize the features. 
We experiment with two datasets and compare our approach with the state-of-the-art method in prescriptive process monitoring.

\subsection{Datasets}
We performed our experiments on two publicly available datasets, namely BPIC12 and BPIC17. We chose these two datasets because, to the best of our knowledge, they are the only publicly available datasets with a treatment present in the log that can affect the process outcome. Both these logs contain traces of a loan origination process. We consider the process outcome to be positive if the customer accepts the loan offer. Also, usually one loan offer is made to each customer. But we observe cases where more than one offer is made to the same customer. We observe that the rate of success is higher for such cases. Hence, we consider multiple loan offers to one customer as a possible treatment. Although these two logs refer to the same process, they have some differences. The BPIC17 log has a considerably larger number of cases and contains more features. 

\subsection{Performance measure}
\begin{figure}[!h]
	\vspace{-\baselineskip}
    \centering
    \vspace{-5mm}
	\subfloat[BPIC 2012]{\label{fig:bpic12}\includegraphics[width=0.52\textwidth, height=4.5cm]{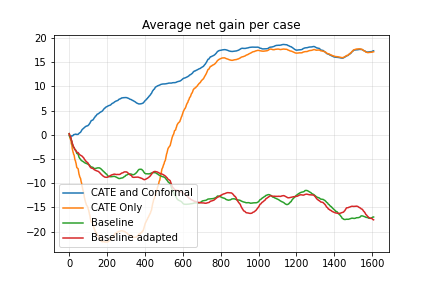}}
	\subfloat[BPIC 2017]{\label{fig:bpic17}\includegraphics[width=0.52\textwidth, height=4.5cm]{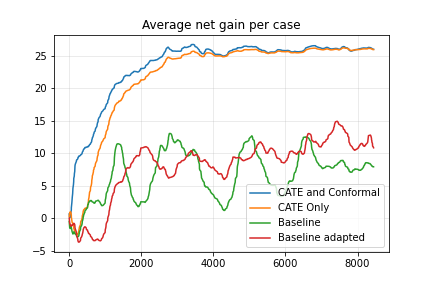}}
	\caption{Average net-gain of cases.}
	\label{fig:netGain}
    \vspace{-5mm}
\end{figure}

In this experiment we evaluate the success of each component of our approach by measuring the following gain function:

\begin{equation}
    NetGain=Y(t)*gain - t*cost,
\end{equation}
where $t\in \{0,1\}$ is the treatment option the agent recommended. The net-gain measures the amount of money that we gain from each case. Negative net-gain represents loss. We first run our experiment only giving the CATE estimates $\theta_{u,k}$ and $\theta_{l,k}$ to the agent. In the second experiment, we give both the CATE estimates and the conformal prediction scores $\rho_k$ to the reinforcement learning agent. We also compare our approach with the one proposed in~\cite{metzger2020triggering} as it is state-of-the-art in addressing the when-to-treat problem. Figure~\ref{fig:netGain} describes the results for both datasets. 

In the figures above, we consider an expensive treatment. Specifically, we suppose that the benefit of a positive outcome is \$50 and the cost of the treatment is \$25. We also experimented with cheap treatments (e.g., \$1). But since cheap treatments do not require a strict policy, we include those results in the supplementary materials. The x-axis in Figure~\ref{fig:netGain} refers to the cases the agent decides on, and the y-axis is the net gain for each case. We can see that the agents using the CATE upper and lower bounds outperform the agents that use predictions about the case outcome. This result is because the agents using predictions, target cases that end up with a bad outcome. But the agents using CATE have information about the effectiveness of the treatment. They only treat if they are confident it will turn a negative outcome into a positive one. In other words, they target the persuadables in Figure~\ref{fig:uplift}. Also, we observe that the agent using both conformal prediction score and CATE estimates, converges faster, empirically proving our hypothesis that conformal prediction can detect the sure things and the lost causes, thereby pruning the search space for the agent and helping it make fewer mistakes. This is important if RL is applied online in a real environment because, without conformal prediction, we observe that the net gain is negative for the first few hundred cases.

We ran two versions of the baseline. The first (shown in green) uses the reward function proposed in~\cite{metzger2020triggering}. In the second version (shown in red), we adapt the baseline to use our proposed reward function (Table~\ref{table:rewards}). We can see that in the BPIC17 dataset, the baseline is unstable. This is because in some cases, the predictive information that the agent has matches the effectiveness of the treatment, so the agent correctly decides to treat. But other times, it treats when the treatment is needed but ineffective. But when using our proposed reward function, we see a slow improvement in performance. This is because the agent is slowly learning to be more selective about when to apply the treatment through the reward function. But the description of its environment (the prediction and its reliability) does not have enough signal to converge faster. In the BPIC12 dataset, both versions of the baseline produce negative net gain. This is because this log is considerably smaller than the BPIC17 log. So the agent does not see enough samples to discover a policy producing a positive net gain. 

We also experimented with the intuitive reward function that does not contain any further punishments for the agent's wrong decisions (see Section~\ref{policyselection}). However, we found that this reward function does not produce good results. This is because this reward function does not have enough signal for the agent to distinguish between correct and incorrect decisions, leading the agent to treat unnecessarily. This problem can be addressed by adjusting the exploration/exploitation rate of the agent. That said, since we are using an algorithm (PPO) that does not allow manual adjustment of this rate, we leave that for future work. The results of this experiment and the specifications of the alternative reward function can be found in the supplementary materials. 

\subsection{Statistical Tests on Enhanced Logs}
In this section, we evaluate the quality of our generated potential outcomes. To this end, we ran a few statistical two-sample tests. We tested the hypothesis that the generated and real outcomes come from different distributions and report the p-values. Any p-value above $0.05$ indicates that the test cannot conclude that the two samples come from different distributions. We use the same tests that we did in~\cite{journalExtension} and report the results in Table~\ref{table:pvalues}. It can be seen that all p-values are above the $0.05$ threshold, meaning that none of the tests can detect that the enhanced and real outcomes come from different distributions.

\begin{table}[]
\centering
\begin{tabular}{|c|c|c|c|c|c|c|c|}
\hline
       & \begin{tabular}[c]{@{}c@{}}Kolmogorov\\ Smirnov\end{tabular} & \begin{tabular}[c]{@{}c@{}}Epps\\ Singleton\end{tabular} & \begin{tabular}[c]{@{}c@{}}Friedman\\ Rafsky\end{tabular} & \begin{tabular}[c]{@{}c@{}}k-Nearest\\ Neighbor\end{tabular} & Energy & \begin{tabular}[c]{@{}c@{}}Wasserstein\\ 1\end{tabular} & \begin{tabular}[c]{@{}c@{}}Wasserstein\\ 2\end{tabular} \\ \hline
BPIC12 & 0.244                                                        & 0.311                                                    & 0.109                                                     & 0.088                                                        & 0.228  & 0.323                                                   & 0.329                                                   \\ \hline
BPIC17 & 1.0                                                          & 0.996                                                    & 0.647                                                     & 0.359                                                        & 0.92   & 0.916                                                   & 0.87                                                    \\ \hline
\end{tabular}
\vspace{2mm}
\caption{Table of p-values for statistical tests}
\label{table:pvalues}
\vspace{-5mm}
\end{table}

\subsection{Threats to Validity}\label{threats}
\vspace{-.3\baselineskip}
The small number of event logs used in our experiments poses threats to external validity. A threat to internal validity is that we generated the two potential outcomes to enhance the logs used in the experiments. The quality of these potential outcomes relies on the quality of the input data. Thus, analysts must ensure that they have a sufficient amount of good-quality data before applying our method. Using observational data to estimate CATEs creates a threat to construct validity. When all possible confounders are not present in the event log, causal models will reduce bias compared to purely correlation-based methods, but they will not eliminate it. The other threat to construct validity is that we use cases as the notion of episodes in the RL phase. This means that if there are long cases in the log, the agent might get exposed to information about the future. To mitigate this threat we did the following. First, we excluded any case information in the description of the environment. The agent only sees CATE estimates, conformal prediction scores, and prefix numbers. However, since these estimates and scores are derived from models using case information, the agent might still be indirectly exposed to future information. Secondly, we sorted the cases by their end timestamp ensuring that as little information about the future is leaked to the agent as possible. 

\section{Conclusion}
\vspace{-\baselineskip}
\label{sec: conclusion}
\vspace{.5\baselineskip}

This paper introduced a method to learn policies to prescribe treatments to cases of a business process in order to maximize the net gain generated by such treatments. The proposed method enhances an existing online RL method by: i) feeding causal effect estimates to obtain a higher net gain by preventing ineffective treatments; ii) using conformal predictions to speed up the convergence of the RL agent; and iii) using a causal dataset enhancement method to simulate an environment where the RL agent can be trained offline.

In the proposed method, each learning episode is one case. This approach potentially exposes the agent to information that is only known in the future, since cases may overlap (data leakage). A natural improvement of this work would be to change the notion of episodes to prevent such leakage. For instance, each episode could be all the events across multiple cases that occur on the same day. The difficulty here is finding a suitable time-step to make good use of the available dataset. Another direction for future work is to experiment with RL algorithms that allow us to manually control the exploration/exploitation rate, giving us enough control for additional tuning. We will also investigate adjusting this rate indirectly, e.g., by adding coefficients to the reward function. Further directions for future work include experimenting with other RL techniques and validating the proposed method via a case study.

\smallskip\noindent\textbf{Reproducibility}
The source code of our tool, the datasets used, and the experiment results can be found at \url{https://github.com/zahradbozorgi/WhenToTreat}

\smallskip\noindent\textbf{Acknowledgments}
Research funded by the Australian Research Council (grant DP180102839), the European Research Council (PIX Project), and the Estonian Research Council (grant PRG1226).
%
%
%
\bibliographystyle{splncs04}
%
\bibliography{7.0.0.library}

\begin{thebibliography}{10}
\providecommand{\url}[1]{\texttt{#1}}
\providecommand{\urlprefix}{URL }
\providecommand{\doi}[1]{https://doi.org/#1}

\bibitem{athey2016recursive}
Athey, S., Imbens, G.: Recursive partitioning for heterogeneous causal effects.
  Proceedings of the National Academy of Sciences  (2016)

\bibitem{DBLP:conf/icpm/BozorgiTDRP20}
Bozorgi, Z.D., Teinemaa, I., Dumas, M., Rosa, M.L., Polyvyanyy, A.: Process
  mining meets causal machine learning: Discovering causal rules from event
  logs. In: 2nd {ICPM} (2020)

\bibitem{DBLP:conf/icpm/BozorgiTDRP21}
Bozorgi, Z.D., Teinemaa, I., Dumas, M., Rosa, M.L., Polyvyanyy, A.:
  Prescriptive process monitoring for cost-aware cycle time reduction. In: 3rd
  {ICPM} (2021)

\bibitem{journalExtension}
Bozorgi, Z.D., Teinemaa, I., Dumas, M., Rosa, M.L., Polyvyanyy, A.:
  Prescriptive process monitoring based on causal effect estimation. Under
  Review  (2022)

\bibitem{dorogush2017catboost}
Dorogush, A.V., Ershov, V., Gulin, A.: Catboost: gradient boosting with
  categorical features support. Workshop on ML Systems at NIPS  (2017)

\bibitem{DBLP:journals/kais/Fahrenkrog-Petersen22}
Fahrenkrog{-}Petersen, S.A., Tax, N., Teinemaa, I., Dumas, M., de~Leoni, M.,
  Maggi, F.M., Weidlich, M.: Fire now, fire later: alarm-based systems for
  prescriptive process monitoring. Knowledge and Information Systems  (2022)

\bibitem{koorn2022mining}
Koorn, J.J., Lu, X., Leopold, H., Martin, N., Verboven, S., Reijers, H.A.:
  Mining statistical relations for better decision making in healthcare
  processes  (2022)

\bibitem{DBLP:conf/bpm/KoornLLR20}
Koorn, J.J., Lu, X., Leopold, H., Reijers, H.A.: Looking for meaning:
  Discovering action-response-effect patterns in business processes. In: {BPM}
  Proceedings (2020)

\bibitem{kubrak2022prescriptive}
Kubrak, K., Milani, F., Nolte, A., Dumas, M.: Prescriptive process monitoring:
  Quo vadis? PeerJ Computer Science  \textbf{8} (2022)

\bibitem{de2020design}
de~Leoni, M., Dees, M., Reulink, L.: Design and evaluation of a process-aware
  recommender system based on prescriptive analytics. In: 2nd {ICPM} (2020)

\bibitem{metzger2020triggering}
Metzger, A., Kley, T., Palm, A.: Triggering proactive business process
  adaptations via online reinforcement learning. In: {BPM} Proceedings (2020)

\bibitem{DBLP:journals/corr/abs-2011-15007}
Neal, B., Huang, C.W., Raghupathi, S.: Realcause: Realistic causal inference
  benchmarking (2021)

\bibitem{DBLP:conf/icml/OprescuSW19}
Oprescu, M., Syrgkanis, V., Wu, Z.S.: Orthogonal random forest for causal
  inference. In: Proceedings of the 36th {ICML} (2019)

\bibitem{padella2022explainable}
Padella, A., de~Leoni, M., Dogan, O., Galanti, R.: Explainable process
  prescriptive analytics  (2022)

\bibitem{DBLP:conf/bpm/QafariA20}
Qafari, M.S., van~der Aalst, W.M.P.: Root cause analysis in process mining
  using structural equation models. In: {BPM} International Workshops, Revised
  Selected Papers (2020)

\bibitem{qafari2021case}
Qafari, M.S., van~der Aalst, W.M.P.: Case level counterfactual reasoning in
  process mining. In: CAiSE Forum Proceedings (2021)

\bibitem{romano2020classification}
Romano, Y., Sesia, M., Candes, E.: Classification with valid and adaptive
  coverage. NeurIPS  (2020)

\bibitem{rubin1974estimating}
Rubin, D.B.: Estimating causal effects of treatments in randomized and
  nonrandomized studies. Journal of educational Psychology

\bibitem{schulman2017proximal}
Schulman, J., Wolski, F., Dhariwal, P., Radford, A., Klimov, O.: Proximal
  policy optimization algorithms. arXiv preprint arXiv:1707.06347  (2017)

\bibitem{sutton2018reinforcement}
Sutton, R.S., Barto, A.G.: Reinforcement learning: An introduction. MIT press
  (2018)

\bibitem{alarmBased}
Teinemaa, I., Tax, N., de~Leoni, M., Dumas, M., Maggi, F.M.: Alarm-based
  prescriptive process monitoring. In: BPM Forum (2018)

\bibitem{wager2018estimation}
Wager, S., Athey, S.: Estimation and inference of heterogeneous treatment
  effects using random forests. Journal of the American Statistical Association
   (2018)

\bibitem{10.1007/978-3-030-58638-6_12}
Weinzierl, S., Dunzer, S., Zilker, S., Matzner, M.: Prescriptive business
  process monitoring for recommending next best actions. In: BPM Forum (2020)

\end{thebibliography}
\end{document}